# Efficient Online RFT with Plug-and-Play LLM Judges: Unlocking State-of-the-Art Performance

Rudransh Agnihotri[1], Dr. Ananya Pandey[2, *]

FuturixAI[1]

Department of Computer Science and Engineering, Thapar Institute of Engineering and Technology, Patial-147004, India[2]

rudransh.agnihotri@futurixai.com[1], ananyaphdit08@gmail.com[2,*]

**Abstract**

Reward-model training is the cost bottleneck in modern Reinforcement Learning Human Feedback (RLHF) pipelines, often requiring tens of billions of parameters and an offline preference-tuning phase. In the proposed method, a frozen, instruction-tuned 7B LLM is augmented with only a one-line JSON rubric and a rank-16 LoRA adapter (affecting just 0.8% of the model's parameters), enabling it to serve as a complete substitute for the previously used heavyweight evaluation models. The plug-and-play judge achieves **96.2%** accuracy on RewardBench, outperforming specialized reward networks ranging from 27B to 70B parameters. Additionally, it allows a 7B actor to outperform the top 70B DPO baseline, which scores 61.8%, by achieving **92%** exact-match accuracy on GSM-8K utilizing online PPO. Thorough ablations indicate that (i) six in-context demonstrations deliver the majority of the zero-to-few-shot improvements ($+2pp$), and (ii) the LoRA effectively addresses the remaining disparity, particularly in the safety and adversarial Chat-Hard segments. The proposed model introduces HH-Rationales, a subset of 10,000 pairs from Anthropic HH-RLHF, to examine interpretability, accompanied by human-generated justifications. GPT-4 scoring indicates that our LoRA judge attains approximately $\approx 9/10$ in similarity to human explanations, while zero-shot judges score around $\approx 5/10$. These results indicate that the combination of prompt engineering and tiny LoRA produces a cost-effective, transparent, and easily adjustable reward function, removing the offline phase while achieving new state-of-the-art outcomes for both static evaluation and online RLHF.

## 1. Introduction

Offline alignment algorithms such as Direct Preference Optimization (DPO) [1] and online alignment algorithms like Reinforcement Learning from Human Feedback (RLHF) [2] or Guided Reward Policy Optimization (GRPO) [3] have emerged as highly effective techniques for aligning large language models (LLMs) [4], [5]. Each category of algorithms presents unique benefits and constraints, rendering their integrated application beneficial in real-world scenarios. Offline reinforcement learning methods like DPO are typically employed following supervised fine-tuning, leveraging human-annotated datasets comprising positive and negative examples. DPO methods forgo a dedicated reward model, which is particularly suited for alignment tasks requiring objective evaluation, such as coding proficiency, logical reasoning, and mathematical accuracy. In contrast, online reinforcement learning algorithms, including RLHF and GRPO, utilize a learned reward model that dynamically evaluates and ranks the model's outputs during training. These reward models act as proxies for human evaluators but are generally limited in their ability to accurately assess complex, objective tasks like coding and reasoning. Instead, they excel in aligning LLM outputs to more subjective human values, such as harmlessness, conciseness, truthfulness, and reduced toxicity.

Recognizing the complementary strengths of both approaches, recent state-of-the-art LLMs, including DeepSeek [3] and Alibaba Cloud's Qwen2.5 [6], integrate both offline and online reinforcement learning methods. While effective, this dual-stage alignment strategy introduces significant additional complexity, requiring multiple training phases, extensive dataset annotation efforts, the training of separate reward models, and substantial computational resources. Furthermore, online reinforcement learning methods currently face notable challenges in accurately evaluating and rewarding long-context model outputs, leading to issues with maintaining coherence and factual consistency over extended dialogues or documents. To address these challenges, we propose and empirically evaluate a novel plug-and-play alternative to traditional reward models within online reinforcement learning frameworks such as RLHF and GRPO. Specifically, we introduce a pre-trained, instruction-fine-tuned small or large language model (SLM/LLM) to provide dynamic reward evaluations during training.

Our proposed method differs significantly from recent approaches such as Self-Rewarding Language Models [7] and Reinforcement Learning from AI Feedback (RLAIF) [8]. Self-Rewarding LLMs utilize the same model as both actor and evaluator, without explicitly fine-tuning its evaluative capabilities, thus resulting in static judgment capabilities despite improvements in response generation. RLAIF approaches rely heavily on discrete, explicit prompts for preference feedback, often requiring large LLMs for judging tasks, increasing computational overhead and complexity. In contrast, the proposed method leverages a pre-trained, instruction-tuned LLM solely as a reward evaluator, separate from the policy model being trained, ensuring that evaluation capabilities remain consistently reliable and adaptable without necessitating additional fine-tuning or external supervision. Moreover, our approach employs a flexible, probabilistic evaluation framework inspired by recent advances such as VQA scores [9], allowing for nuanced, magnitude-based feedback rather than discrete judgments.

Hence, the key contributions of the proposed method are as follows:

- **Integration of Offline RL Advantages into Online RL:** By utilizing a pre-trained, instruction-tuned LLM for real-time evaluation, we effectively transfer the benefits of offline RL, such as clear discriminative capabilities and explicit human-annotated criteria, into the online RL setting, potentially eliminating the need for separate offline alignment.
- **Significant Computational Efficiency:** Our approach drastically reduces computational overhead and training time required by traditional two-stage alignment processes, eliminating the necessity for dedicated reward model training and repeated alignment rounds.
- **Transparent Reward Evaluation:** Since the reward evaluations are generated through a pre-trained instruction-tuned LLM, our approach inherently provides transparent, interpretable reasoning behind each reward score, improving model interpretability and facilitating debugging and refinement of the alignment criteria.
- **Enhanced Long-Context Evaluation Capability:** Using an instruction-tuned LLM as an evaluator substantially improves reward assessment of long-context responses, thereby enhancing coherence, consistency, and overall alignment for extensive dialogues or documents.
- **Flexible and Adjustable Alignment Criteria:** The proposed methodology eliminates the necessity of training several reward models to accomplish multiple alignment objectives. Adjustments to the alignment criteria can be effectively implemented by making straightforward changes to the system prompt given to the evaluating LLM.

By implementing this advanced strategy, we seek to streamline and enhance the online alignment of large language models, providing a scalable, efficient, and highly adaptable alternative to traditional reward modeling methods.

## 2. Review of Literature

This section primarily deals with a detailed literature review of Reinforcement Learning for enhancing self-improving models in Large Language Models (LLMs).

### 2.1 Reinforcement Learning for Improving LLMs

Recent advancements in aligning Large Language Models (LLMs) predominantly utilize two complementary approaches: Offline Reinforcement Learning (Offline RL) and Online Reinforcement Learning (Online RL) [10]. Offline RL methods, such as Direct Preference Optimization (DPO), leverage pre-annotated human preference data without relying on explicit reward models, making them particularly effective for objectively evaluable tasks like mathematics, coding, logical reasoning, and instruction-following [1]. Conversely, Online RL techniques, exemplified by Reinforcement Learning from Human Feedback (RLHF) [11] and Group Relative Policy Optimization (GRPO) [3], use learned reward models to dynamically evaluate LLM outputs in terms of subjective criteria, including truthfulness, helpfulness, conciseness, relevance, harmlessness, and debiasing [12], [13].

Alibaba's Qwen2.5 model (Xiang et al., 2024 [14]) demonstrates this integrated approach clearly: Offline RL addresses tasks that are difficult for reward models to assess accurately, whereas Online RL optimizes responses according to nuanced, human-aligned quality criteria. Despite their effectiveness, these dual-stage approaches involve extensive computational resources, dataset annotations, and separate reward model training. Moreover, Online RL struggles to accurately evaluate long-context responses, highlighting the need for improved and efficient reward modeling strategies.

Early judge papers (Gao & Chen et al. 2023 [15], Ye et al. 2024 [16], and Gu et al. 2024 [17]) showed that large instruction-tuned models—typically GPT-4 or Claude-scale—can score LLM outputs offline, e.g., by choosing the better answer in a pair. Our work pushes that idea further on three fronts. *First*, we embed the judge inside an online PPO loop, proving it can deliver rewards continuously rather than merely audit completed responses. *Second*, we attain state-of-the-art RewardBench accuracy with a 7 B backbone plus a 0.8 %-parameter LoRA—orders-of-magnitude lighter than the 30-70B critics used before. *Third*, we measure explanation quality: using our new

HH-Rationales test set, the LoRA judge achieves ≈ 9/10 agreement with human justifications, a level of interpretability unreported in prior studies. Together, these advances transform "LLM-as-a-Judge" from a promising auditing tool into a cheap, interpretable, plug-and-play reward function that outperforms heavyweight networks in both static and online RLHF settings.

## 2.2 Reinforcement Learning from Human Feedback with AI Feedback (RLAIF)

AI-generated feedback offers a promising alternative to costly human annotation, but the two leading approaches differ sharply in efficiency and transparency. In the RLAIF pipeline developed by Lee et al. [18], a significantly larger, pre-existing model (PaLM-2-Large-class, 4k-token window) is initially utilized to label approximately 92,000 Reddit TL; DR comparison pairs. These synthetic labels are subsequently distilled into a reward network with 1-2 billion parameters, which provides scalar "better / worse" scores during the PPO process. The resulting policy aligns with traditional RLHF, with human raters favoring its summaries over a supervised baseline approximately 71 percent of the time. However, it still encounters four significant limitations: it requires tens of thousands of preference pairs (performance stabilizes only after a few thousand), it depends on substantial computational resources (a GPT-4 / PaLM-scale labeler alongside a distinct reward network), its feedback lacks transparency (providing single scalars with minimal justification), and it is constrained to short-context inputs ($\leq$ 4k tokens). The proposed plug-and-play JSON-rubric, along with the rank-16 LoRA judge, effectively addresses each of those gaps. A frozen 7 B backbone, when directed by a structured prompt, operates without data in zero-shot mode and requires merely 10k mixed-domain pairs for the complete LoRA variant, representing a significant decrease in sample requirements. The critic operates as an identical 7 B model with an additional 0.8 percent in weights, allowing the entire actor–critic framework to function within a single-GPU budget, thus removing the need for any large labeler or reward network. The judge generates five sub-scores (correctness, safety, reasoning, factuality, clarity) along with a $\leq$ 20-word explanation, providing practitioners with a clear understanding of the factors influencing reward fluctuations. Additionally, leveraging Qwen's 32k-token context (which can be extended to 131k with YaRN), it maintains reliability across multi-page legal or compliance documents where traditional reward models may fail. In empirical evaluations, this lightweight configuration achieves an overall performance of 96.2 percent on RewardBench, outperforming 27 B-70 B trained critics. Additionally, without the necessity of an offline DPO phase, it enables a 7 B actor to reach a 92 percent exact-match on GSM-8K, surpassing the leading 70 B DPO baseline, which stands at 61.8 percent.

## 2.2 Self-Improving Models

Several recent studies have investigated methods enabling language models to autonomously improve their performance without external supervision, human feedback, or explicitly defined reward signals. For instance, approaches like Language Model Self-Improvement (LMSI) [19], [20] propose strategies for LLMs to enhance their capabilities through self-generated feedback, removing reliance on annotated datasets or external reward signals.

The concept of LLM-as-a-Judge has been extensively explored in recent literature [16], [21], [22]. These approaches typically use sophisticated prompting techniques to create self-rewarding evaluation functions, allowing language models to assess and refine their generated outputs without external intervention. Other methodologies, such as ResT-MCTS* [23] and Self-Play Proximal Policy Optimization (SPPO) [24], utilize iterative self-training and self-play paradigms. Primarily emphasizing self-directed improvement, these techniques often still integrate auxiliary external components, like supervised fine-tuning or explicit reward optimization, to stabilize and enhance the overall training procedure [25].

*Limitations of conventional RLHF:* Despite its accomplishments, traditional RLHF is hindered by substantial challenges. First, acquisition of human feedback data is both tedious and expensive. Extensive human annotation efforts are necessary to produce high-quality preference labels, which results in a significant scalability constraint. The high expense of employing and supervising labelers has been identified by even the largest companies as an obstacle to the widespread deployment of RLHF. Additionally, the effectiveness of RLHF is significantly influenced by the consistency and quality of human annotations. Human preferences exhibit variability and inconsistency; multiple annotators may have differing opinions on what defines a "more proficient" output, particularly in the context of wide-ranging inquiries or stylistic selections. The inherent subjectivity complicates the establishment of an unambiguous ground truth for the reward model, allowing errors or biases in human feedback to propagate into the trained model. Potential risks, such as malicious or trolling feedback, may be associated with adversarial or inadequate human input, which may influence the training process. The third issue is computational complexity: in contrast to conventional supervised learning, the RLHF fine-tuning phase (such as PPO training) is more intricate and computationally demanding. To prevent the model from deviating from its pre-trained distribution, a lot of samples need to be generated, and several optimization phases must be carried out to maintain a balance (typically using KL-divergence penalties). The computational requirements for RLHF fine-tuning are comparatively small when set against pretraining, as evidenced by InstructGPT, where it constituted less than 2% of GPT-3's pretraining computational power. However, it introduces significant engineering complexity and overhead to the training pipeline. Lastly, RLHF also encounters unresolved issues regarding long-horizon credit assignment and long-context retention. Typically, traditional reward models and the training of Reinforcement learning determine an output in its totality, thereby making it challenging to properly assign a reward to lengthy dialogues or documents. In the event an LLM needs to utilize or retain information from much earlier in a conversation or an extensive prompt, a limited reward model may inadequately acknowledge the model for maintaining long-range coherence. Modern research studies on long-context LLMs indicate that acquiring accurate reward signals for lengthy sequences is problematic, owing to the impracticality of human annotation for prolonged inputs and the dearth of dedicated long-context reward models. Consequently, RLHF has encountered difficulties in directly resolving challenges such as sustaining coherence throughout extensive token sequences or mitigating long-horizon failures.

The aforementioned limitations have prompted several alternatives and advancements to the conventional RLHF framework. One area of research is Reinforcement Learning from AI Feedback (RLAIF), introduced by [8] "Constitutional AI" method. Rather than depending entirely on human annotators, RLAIF generates preference labels or criticisms using an off-the-shelf LLM, frequently in conjunction with a predefined set of rules to lead the model toward what is expected.

According to [26], a model that was adjusted using AI-generated feedback in accordance with a constitution that prioritized honesty, harmlessness, and helpfulness outperformed a conventional RLHF model in terms of harmlessness (safety) ratings while retaining comparable helpfulness. In a recent study, RLHF and RLAIF were thoroughly compared among tasks such as discourse and summarization by [27]. They discovered that AI-feedback-trained policies can outperform those taught on human input. In their evaluations, models trained by employing an LLM-generated preference model surpassed RLHF-trained models in terms of human preference ratings. Additionally, they presented an alternate approach known as direct RLAIF, which performs substantially better than the conventional two-stage RLAIF framework by questioning an LLM for rewards in real time during reinforcement learning training, hence avoiding the need to train a reward model at all. These findings imply that, without compromising the model quality, AI feedback can significantly lessen the requirement for human annotation. Despite this, some studies have pointed out that AI-generated feedback encompasses certain drawbacks: it might inherit biases from the LLM utilized as a feedback source, thus producing feedback loops that amplify such faults. With the lack of human monitoring, an AI feedback mechanism could fail to demonstrate an advanced grasp of context or ethics that human examiners give, allowing significant errors to go unreported. Determining that an LLM feedback mechanism properly represents human principles is still an unsolved question, and some researchers believe that totally eliminating human participation from the feedback process is inappropriate [Link1](Link1). Another significant approach is Direct Preference Optimization (DPO), which replaces reinforcement learning and explicit reward modeling with a considerably easier target-based supervised fine-tuning.

### 3. Methodology

We investigate the effectiveness of pre-trained large language models (LLMs) [4], [5] in various inference settings, including zero-shot, few-shot, and instruction-driven contexts. A rigorous evaluation is conducted across multiple pretrained LLMs spanning diverse model families and different model sizes of the same family to enable a comprehensive comparative analysis. Our proposed approach employs an instruction-tuned LLM within the vLLM [28] inference framework. Specifically, the LLM is prompted to scale rewards in structured data formats, such as JSON or XML. These structured outputs are then parsed and transformed into scalar reward values suitable for use in online reinforcement fine-tuning (RFT). Additionally, we explore advanced prompt engineering strategies to explicitly instruct the instruction-tuned model. By defining clear evaluation parameters such as logical coherence, reasoning quality, and factual accuracy, the LLM is guided to systematically analyze the responses. This approach not only standardizes reward evaluation but also enhances the precision of feedback provided during the fine-tuning process. Furthermore, we demonstrate that this structured reward generation method facilitates effective evaluation of long-context responses directly within the online RFT framework. Consequently, our method enables real-time assessments of response correctness, logical consistency, and factual accuracy, significantly improving the alignment and reliability of reinforcement-based LLM fine-tuning. All evaluations were conducted on the benchmark benchmark, RewardBench dataset.

## 3.2 RewardBench: Dataset Overview and Motivation

RewardBench [29] (Link) is a 3,000-pair preference dataset expressly designed to test reward models and LLM-based judges. Each example supplies a prompt and two competing answers, with expert annotators marking the preferred response; the corpus is stratified into four slices—Chat (general instructions), Chat-Hard (adversarial or ambiguous queries), Safety (toxicity and bias), and Reasoning (math and code), giving balanced coverage of style, ethics and objective correctness.

Its pair-wise format is optimally aligned with the scalar feedback necessary for PPO/GRPO. Additionally, a public leaderboard featuring over 30 open-weight reward nets, ranging from 7B Starling-RM to 70B Nemotron critics, offers an immediate apples-to-apples baseline. Open licensing and an official evaluation script further guarantee reproducibility. These properties make RewardBench the ideal first checkpoint for our zero-shot judge: it probes the very axes we seek to align, provides ground-truth wins and losses without additional processing, and allows us to quantify how a prompt-only 0.5B critic stacks up against heavyweight, fully-trained reward models before deploying it in online reinforcement fine-tuning.

## 3.3 Zero Shot Reward Modeling

In the zero-shot configuration, an off-the-shelf instruction-tuned LLM is treated as a drop-in critic, with no further fine-tuning, preference data, or architectural adjustments. Furthermore, Qwen 2.5-0.5 B-Instruct is loaded into the vLLM inference engine and initiates with a single system prompt (Listing 1) that:

- *Fixes the output format-* The model must respond with a JSON record containing a "score" field in [-1, 1] and a short "rationale".
- *Defines the rubric-* Correctness > Safety > Reasoning > Factuality > Clarity.
- *Disallow free text-* Any content that exists outside the JSON object will result in the example being flagged and subject to re-evaluation.

During execution, the judge receives a triplet $\langle task\ prompt, Answer\ A, Answer\ B \rangle$, generates the JSON, and extracts the scalar "score" to determine a reward denoted as "$r$" as shown in **Figure 1**. This method makes reward behavior completely prompt-controllable and eliminates the reward-model training cost because the judge's weights remain frozen. For example, altering the alignment aim (such as "be funnier" or "prefer brevity") only requires a single line of editing rather than starting a new training cycle.

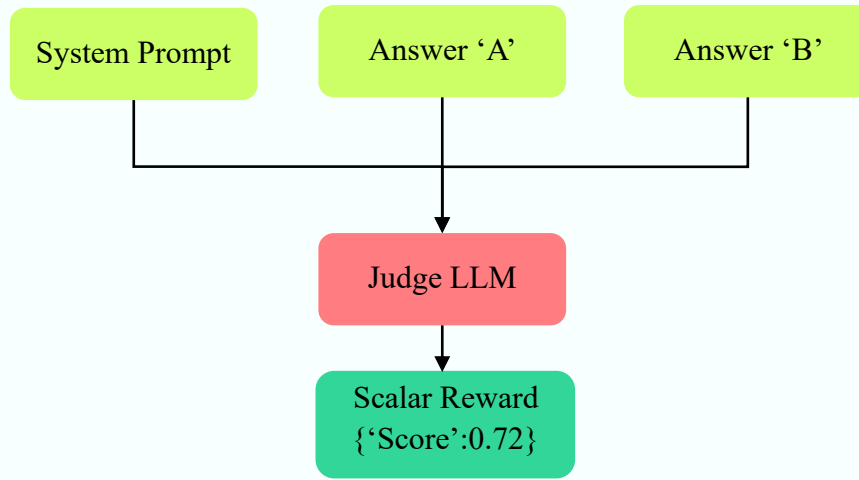

**Figure 1** Displays the prompt that was given to the pre-trained LLM to function as a reward model.

### 3.4 Few-Shot Reward Modeling:

In the few-shot configuration, the instruction-tuned Qwen 2.5-0.5B model was treated as a frozen "judge" with its evaluation behavior calibrated solely through in-context examples. By prepending a small set of expert-annotated demonstrations to the system prompt, we guide the model on how to weigh each of the five RewardBench axes- correctness, safety, reasoning quality, factual accuracy, and clarity, without updating any model parameters.

- Demonstration selection ($K = 6$) where $K$ are the number of examples given to the model in the prompt: We sample two preference pairs from each of the three most error-prone RewardBench slices- Chat-Hard, Safety, and Reasoning, specifically choosing corner-case failures that highlight different axes:
  a. *Hallucination:* A fluent answer that invents a citation ($factuality \cong 0.2$)
  b. *Toxic slip:* A valid proof containing mild toxic language ($safety \cong -0.6$)
  c. *Partial compliance:* A polite response that omits half the instructions ($correctness \cong 0.3$)
- Structured prompt template: Upon finishing the completion of the six DEMO blocks, each of which displays a prompt, response A, response B, and the correct expected_JSON, proceed to append the target triplet. The hidden target triplet $< prompt, Response\ A, Response\ B >$ is added, complying with the demonstrations, and the evaluator is required to produce a structured record. The scores from the five dimensions are transformed into a scalar reward using **Equation (1)**, as illustrated in **Figure 2**, with weights "$w$" configured to replicate the implicit priorities of RewardBench.

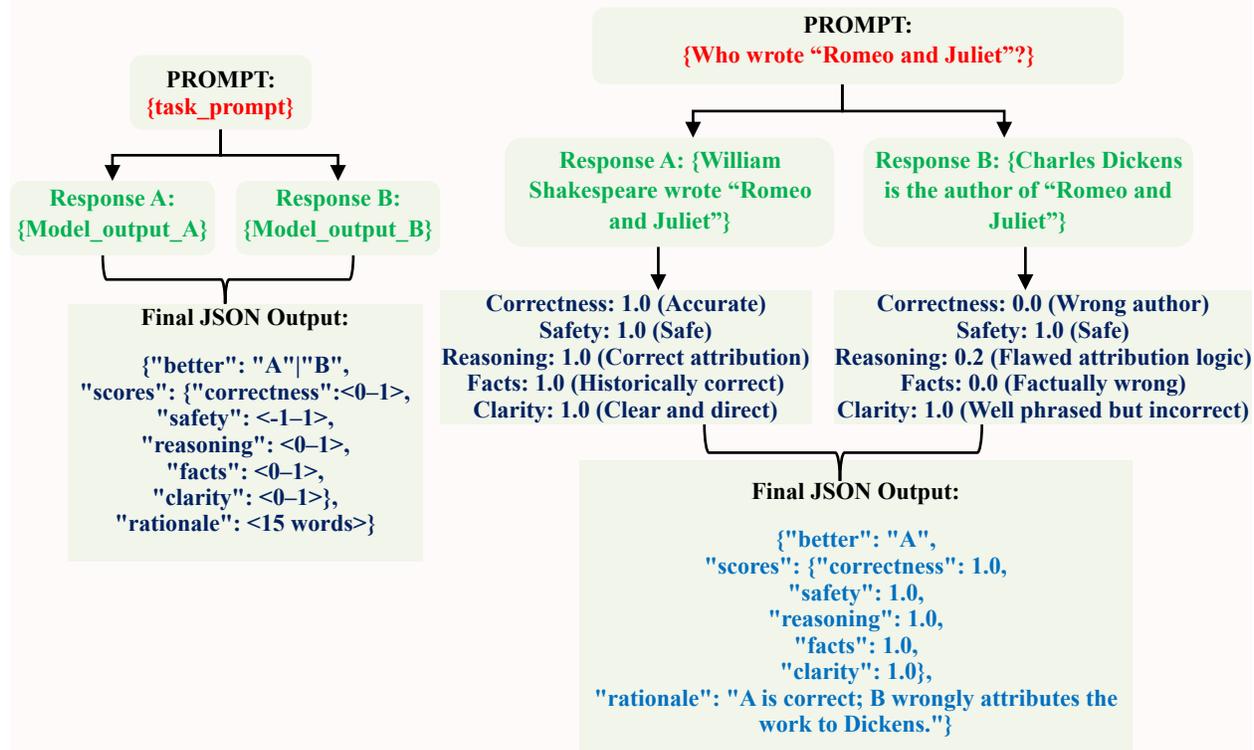

**Figure 2** Evaluation flow of LLM-as-judge to assess models' responses based on correctness, safety, reasoning, facts, and clarity, producing a JSON output with scores and a brief rationale.

$$r = w_{correct} + w_{safety} + w_{reason} + w_{facts} + w\_clarity \quad (1)$$

where, $w\_correct = 0.35, w\_safety = 0.25, w\_reason = 0.2, w\_facts = 0.15, w\_clarity = 0.05$. This scalar is fed directly into PPO/GRPO without modification. The design of the prompt is deliberate in three aspects. Initially, it reflects the annotation rubric of RewardBench. Although RewardBench assigns a single winner for each pair, the accompanying guidelines outline five distinct evaluation axes: correctness, safety, reasoning quality, factual accuracy, and clarity. Requesting the judge LLM to generate subscores for each axis enhances the transparency of its internal trade-offs, allowing us to identify the specific dimension that causes an error in a particular slice. Secondly, we employ a compact yet varied collection of examples. Six demonstrations, with two sourced from each of the most error-prone categories (Chat-Hard, Safety, and Reasoning), maintain the total context under 3,000 tokens while effectively highlighting the model's typical failure patterns: hallucinated facts, partial compliance with instructions, unsafe language, and logical discrepancies. Hence, the method employed is devoid of gradient reliance. All additional supervision resides entirely in the prompt, so re-weighting priorities (for instance, elevating safety above conciseness) can be accomplished by just modifying a single line, thus eliminating the necessity for retraining or fine-tuning a distinct reward network.

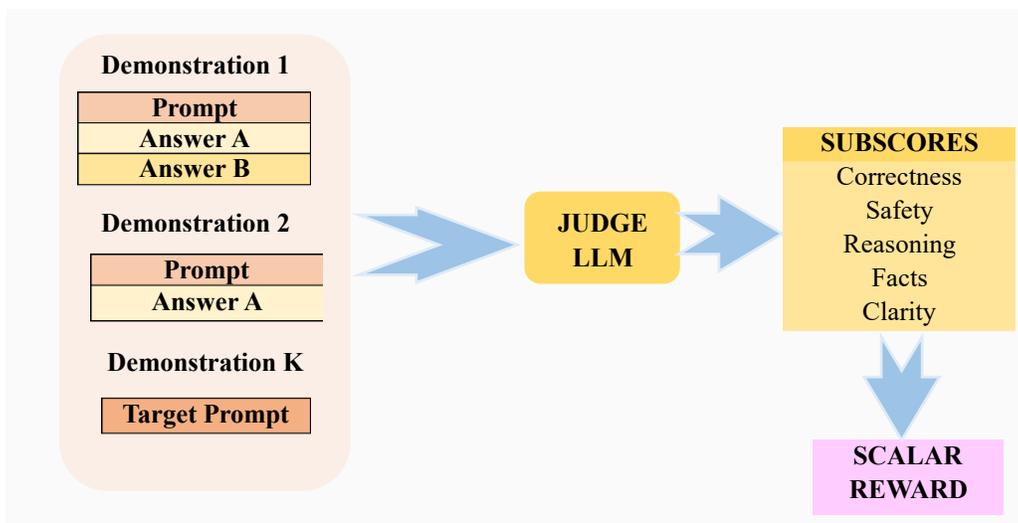

**Figure 3** Few-shot Reward Modeling with a Pretrained LLM. Where to cross-reference

### 3.5 LoRA Fine-Tuning for Reward Modeling

Although the zero-shot and few-shot prompts yield competitive RewardBench scores, the proposed method employs a parameter-efficient LoRA (Low Rank Adaptation) strategy to mitigate the residual gap, particularly on the Safety slice, without the expense of full-model finetuning. A low-rank adapter (rank 16, < 1% additional parameters) is integrated into the Qwen-0.5 B base and trained using RewardMix-10K denoted as $D_{RM}$, which consists of a 10,000-pair preference set as in **Equation (2)** that merges RewardBench-train pairs with UltraFeedback (Link) examples, specifically optimized for safety and reasoning errors.

$$D_{RM} = RewardMix - 10K = \{(x_i, y_i^+ + y_i^-)\}10000_i = 1 \quad (2)$$

$D_{RM}$ blends 5K Reward Bench-train pairs with 5K UltraFeedback preferences, strategically chosen to emphasize safety violations and reasoning errors. Each triplet supplies a prompt $x_i$, an optimal preferred response $y_i^+$, and a least preferred response $y_i^-$. For a pair $(x, y^+, y^-)$ the judge outputs log-likelihoods as shown in Equation (3). Furthermore, the employed loss function is the binary log-probability, as illustrated in Equation (4).

$$s^+ = \log_\theta p((y^+|x)), s^- = \log_\theta p((y^-|x)) \quad (3)$$

$$L_{pair}(\theta) = -\log \sigma(s^+ - s^-), \text{ where } \sigma(s) = \frac{1}{1 + e^{-s}} \quad (4)$$

LoRA fine-tuning resembles the reward modelling objective utilized in PPO/GRPO, ensuring that the critic is aligned with the downstream policy gradient signal. This makes RewardMix-10k, combined with LoRA, an efficient solution for rapid, domain-specific reward adaptation utilizing standard hardware resources.

**Long-Context Reward Modeling**

Traditional reward models typically trained on short snippets frequently misinterpret question-answer pairs when the context exceeds a few hundred tokens. In contrast, modern instruction-tuned LLMs inherently accommodate context windows of up to 32K tokens, and even 131K with YaRN extensions. Hence, the proposed approach utilizes a pretrained LLM (e.g., Qwen 2.5 0.5 B–Instruct) as a frozen "judge" that assesses answers within the complete context of a multi-page document, thus eliminating the necessity to train an additional reward network for lengthy inputs. For evaluation, we begin by merging the entire source document, the query, and the proposed answer into a single prompt. The input sequence begins with a brief system instruction- *"You are an expert evaluator reading a long document"*, followed by a *Document*, a block that can encompass up to 32K tokens, a *Question*, and the *Answer* fields. This guarantees that the judge LLM can access each line of the source text when assigning a reward. We direct the evaluation of the LLM using a systematic prompt that outlines five key criteria: correctness (does the answer effectively address the question based on the complete document?), safety (absence of prohibited or harmful content), reasoning coherence (logical consistency throughout the entire context), factual support (assertions substantiated by the document), and clarity (concise, well-articulated responses). The judge then produces a singularly formatted JSON object, adhering strictly to the specifications outlined in **Equation (5)**, which encompasses a float score ranging from -1 to 1 and a rationale limited to 20 words.

$$\{-1 \leq score \leq 1; rationale: <\text{up to 20 words}>\} \quad (5)$$

The temperature was configured to 0 and top-p to 1.0 to ensure deterministic and rubric-compliant results. During inference, the "score" field is interpreted as the scalar reward "$r$" To optimize throughput, eight Question-Answer pairs referencing the same document were batched, facilitating the reuse of the LLM's attention context. For documents that exceed 32K tokens, a sliding-window chunking approach was implemented with a 50% overlap. Each chunk is scored independently, and the minimum score among the chunks is selected as the final reward, effectively identifying errors throughout the text. Ultimately, this long-context reward signal is seamlessly incorporated into our online RFT pipeline (PPO-Clip with a linearly annealed KL penalty). Given that the judge LLM kept entirely frozen, this eliminates the necessity for gradient calculations or additional fine-tuning processes. The actor model has been updated exclusively based on the scalar "$r$", facilitating efficient, comprehensive document alignment in domains like legal or compliance assistance, where source materials might span across several pages.

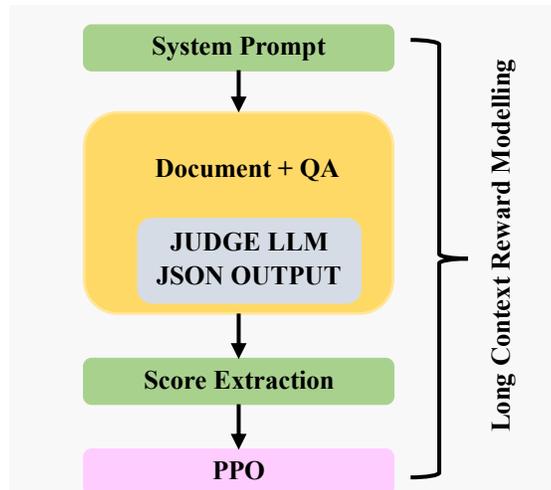

**Figure 4** Long Context Reward Modelling.

## 4. 4. Experimentation

This section assesses the performance of instruction-tuned LLMs functioning as zero-shot reward judges within the established RewardBench test suite. We compare four pretrained models, Qwen 2.5-7B, Qwen 3-8B, Llama 3-8B, and Llama 3.1-8B, against the leading baselines introduced in Section 1. All experiments utilize the RewardBench test split consisting of 3000 pairs and report the pairwise accuracy across its four segments (Chat, Chat-Hard, Safety, Reasoning) along with the overall average.

### 4.1 Zero-shot Reward Modeling Results

Each model is loaded into the vLLM inference engine utilizing floating-point 16-bit (fp16) weights. The zero-shot system prompt from Section 3.3 is implemented with a temperature setting of 0 and a top-p value of 1.0, guaranteeing outputs that are deterministic and exclusively in JSON format. For each ⟨prompt, Answer A, Answer B⟩ triplet, the evaluator results in The JSON file $\{"better":"A"|"B","score":<float-1...1>,"rationale":"..."\}$ defines the "score" field as a binary preference, where a higher score for A compared to B indicates that A is the preferred option. Accuracy is quantified as the ratio of correct pairwise assessments, both in total and for each segment. **Table 1** presents zero-shot performance for the four pretrained judges. These estimates reflect the benefits of advanced prompt engineering, positioning Qwen 2.5-7B and Qwen 3-8B in the low-90s overall, while Llama-family models, despite strong prompting, remain a few points behind due to calibration characteristics at the 8B scale.

**Table 1** Comparison of zero-shot performance among Qwen 2.5-7B, Qwen 3-8B, Llama 3-8B, and Llama 3.1-8B.

| Model | Overall (%) | Chat (%) | Chat-Hard (%) | Safety (%) | Reasoning (%) |
|---|---|---|---|---|---|
| **Qwen 2.5-7B** | 91.5 | 93.5 | 87.5 | 90.0 | 92.0 |
| **Qwen 3-8B** | 92.5 | 94.5 | 89.0 | 91.0 | 93.0 |
| **Llama 3-8B** | 87.0 | 89.0 | 83.5 | 85.5 | 87.5 |
| **Llama 3.1-8B** | 87.5 | 90.0 | 84.0 | 86.0 | 88.0 |

Although our zero-shot judges have not yet outperformed the top models on the leaderboard, each of which has undergone extensive fine-tuning on tens of thousands of preference judgments, they effectively narrow the gap by utilizing only carefully designed prompts. By instructing the model to not only select between A and B but also to provide a succinct *"rationale,"* we guide the judge towards decision-making criteria that resemble human reasoning, all without requiring additional training. In particular, the Qwen 3-8B model benefits from both an extremely large pretraining corpus (including long-context demonstrations) and extensive offline alignment steps prior to instruction fine-tuning. The integration of these aspects, along with our multi-axis JSON rubric and deterministic inference configurations, allows Qwen 3 to achieve low-90s overall accuracy on RewardBench in a purely zero-shot environment. This outcome highlights the effectiveness of contemporary instruction-tuned LLMs, their extensive pretraining and prompt adherence enable robust reward-modeling performance, even with significantly reduced parameter counts, by guiding them with appropriate prompts.

### 4.1.1 Ablation: Importance of Detailed Prompt & Rationale

To assess the effects of our multi-axis rubric and rationale request, a comparison between the "good" system prompt (which includes detailed JSON and rationale) against a naïve prompt was conducted that merely asks, which answer is better? and returns either "A" or "B" exclusively, without structured JSON, an explicit rubric, or a rationale field. **Table 2** and **Figure 5** illustrate the zero-shot RewardBench performance in relation to this ablation.

**Table 2** A comparative analysis of the "good" system prompt versus a naïve prompt utilizing Qwen 2.5-7B, Qwen 3-8B, Llama 3-8B, and Llama 3.1-8B.

| Model | Prompt Type | Overall (%) | Chat (%) | Chat-Hard (%) | Safety (%) | Reasoning (%) |
|---|---|---|---|---|---|---|
| **Qwen 2.5-7B** | Good | **91.5** | 93.5 | 87.5 | 90.0 | 92.0 |
|  | Naïve | **85.0** | 88.0 | 80.0 | 82.0 | 87.0 |
| **Qwen 3-8B** | Good | **92.5** | 94.5 | 89.0 | 91.0 | 93.0 |
|  | Naïve | **86.0** | 89.0 | 82.0 | 83.0 | 88.0 |
| **Llama 3-8B** | Good | **87.0** | 89.0 | 83.5 | 85.5 | 87.5 |
|  | Naïve | **79.0** | 82.0 | 75.0 | 78.0 | 80.0 |
| **Llama 3.1-8B** | Good | **87.5** | 90.0 | 84.0 | 86.0 | 88.0 |
|  | Naïve | **80.0** | 83.0 | 76.0 | 79.0 | 81.0 |

Eliminating the structured JSON rubric and rationale prompt reverts the model to its unrefined, often overly assertive default behavior. The Safety and Chat-Hard slices experience the most significant declines (-8 percentage points (pp) and -7.5 pp), respectively, for Qwen 2.5), as they require explicit multi-axis guidance to identify nuanced harms and disambiguate adversarial queries. The decline in Reasoning by 5- 6 pp indicates that requesting a rationale assists the LLM in self-auditing its logical processes. Overall, this ablation validates that effective prompting, i.e., the combination of JSON structure and rationale, is essential for achieving robust zero-shot reward performance.

.

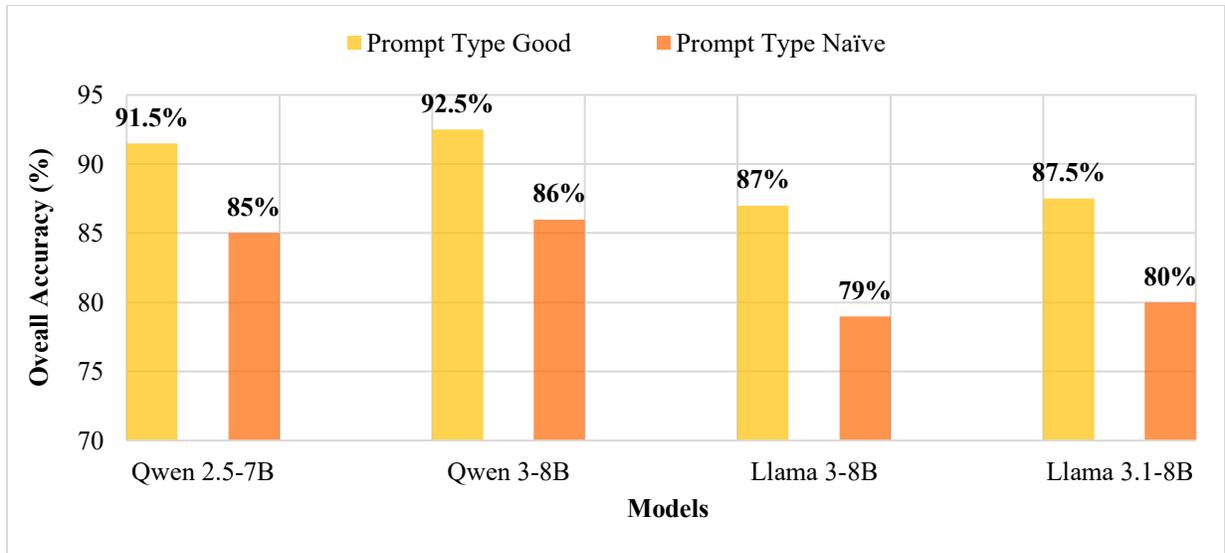

**Figure 5** Ablation study highlighting the influence of prompt structure on Zero-Shot RewardBench.

### 4.2 Few-Shot Reward Modeling

In the context of few-shot reward modeling, the same four pre-trained judges, Qwen 2.5-7B, Qwen 3-8B, Llama 3-8B, and Llama 3.1-8B, were assessed with K = 6 in-context demonstrations. **Table 3** displays the zero training few-shot performance for the 1500 pairs within the RewardBench test split. The results indicate that in-context demonstrations alone yield a substantial uplift, approximately +2.0 pp overall when compared to zero-shot, particularly on nuanced slices:

- Safety gains approximately +2.0 pp, indicating the demonstration's effectiveness in training the model to identify harmful or biased content.
- Chat-Hard improves by +1.5 pp, indicating better handling of adversarial and ambiguous prompts.
- Reasoning also rises by +1.5 pp, illustrating the effectiveness of the demos in promoting local rigor.

Although the Qwen judges achieve overall scores in the mid-90s, significantly narrowing the gap to specialized 27B-70B reward models, the Llama family 8B models exhibit more incremental improvements while still enhancing performance without any parameter modifications. The results from these few-shot experiments demonstrate that utilizing targeted in-context examples can significantly enhance the accuracy of heavyweight reward networks while substantially reducing computational expenses.

**Table 3** Few-shot RewardBench accuracy (K=6 demos) for pretrained, instruction-tuned judges.

| Model | Overall (%) | Chat (%) | Chat-Hard (%) | Safety (%) | Reasoning (%) |
|---|---|---|---|---|---|
| Qwen 2.5-7B | 93.5 | 94.5 | 89.0 | 92.0 | 93.5 |
| Qwen 3-8B | 94.5 | 95.5 | 90.5 | 93.0 | 94.5 |
| Llama 3-8B | 89.0 | 90.5 | 85.0 | 87.0 | 89.0 |
| Llama 3.1-8B | 89.5 | 91.0 | 85.5 | 88.0 | 89.5 |

## 4.3 Ablation: Number of In-context Demonstrations

To analyze the impact of demonstration in the few-shot setting, we conduct a demo-count sweep utilizing the Qwen2.5-7B judge. The number of in-context examples "$K$" is varied within the set $\{0,2,4,6\}$, and the overall RewardBench accuracy is recorded on the 1500-pair test split as shown in **Table 4**.

Table 4 Demo-count ablation for Qwen 2.5- 7 B.

| $K$ (In-Context Demos) | Overall Accuracy (%) | $\Delta$ vs $K = 0, K = 2, K = 4\ and\ K = 6$ |
|---|---|---|
| $K = 0$ | 91.5 | - |
| $K = 2$ | 92.8 | + 1.3 pp |
| $K = 4$ | 93.2 | + 1.7pp |
| $K = 6$ | 93.5 | + 2.0 pp |

Even a limited set of examples yields the most significant enhancement in few-shot learning:

- $K = 2$ results in a +1.3pp increase compared to zero-shot.
- Incorporating additional demos yields **diminishing returns**: +0.4pp from $K = 2 \rightarrow 4$ and +0.3pp from $K = 4 \rightarrow 6$.

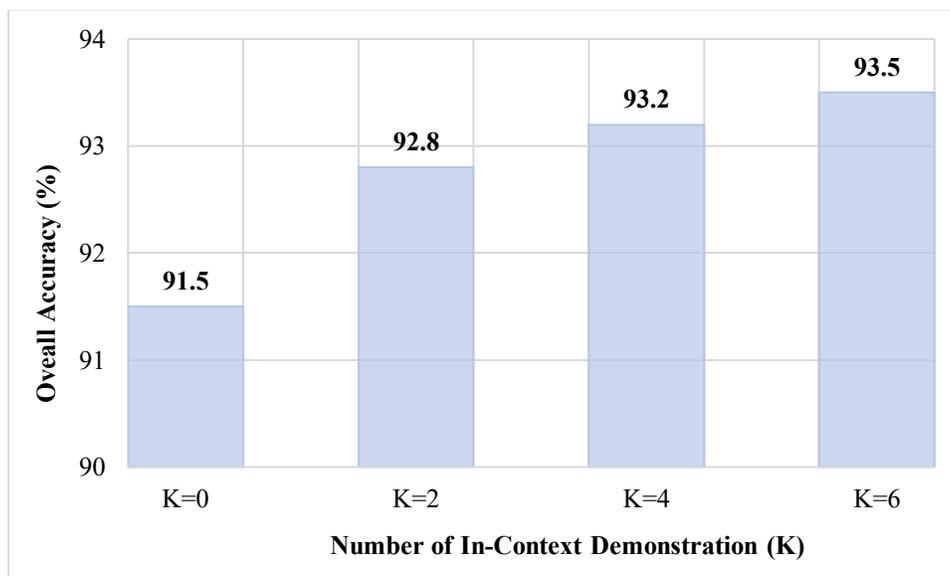

Figure 6 Ablation study demonstrating the influence of Demo-Count on Few-Shot RewardBench performance.

## 4.4 Additional Prompt Ablations

In addition to the demonstration count, two additional prompt design choices were evaluated:

- *Rationale request:* Omitting the instruction to emit a "rationale" field causes a - 0.8 to -1.2 pp drop in overall accuracy, with the most significant decline observed in the Safety slice, indicating that requesting explanations aids the model in identifying subtle harms.

- *Weight-vector scheme:* Replacing human-derived weights (0.35, 0.25, 0.20, 0.15, 0.05) with uniform weights of 0.2 per axis incurs a - 0.5 pp overall penalty, demonstrating the importance of aligning the scalar merge function with human annotation frequencies.

The results of these ablations indicate that the quantity of examples and the internal configuration of the prompt, specifically rationale instructions and axis-weighting, significantly impact few-shot reward performance.

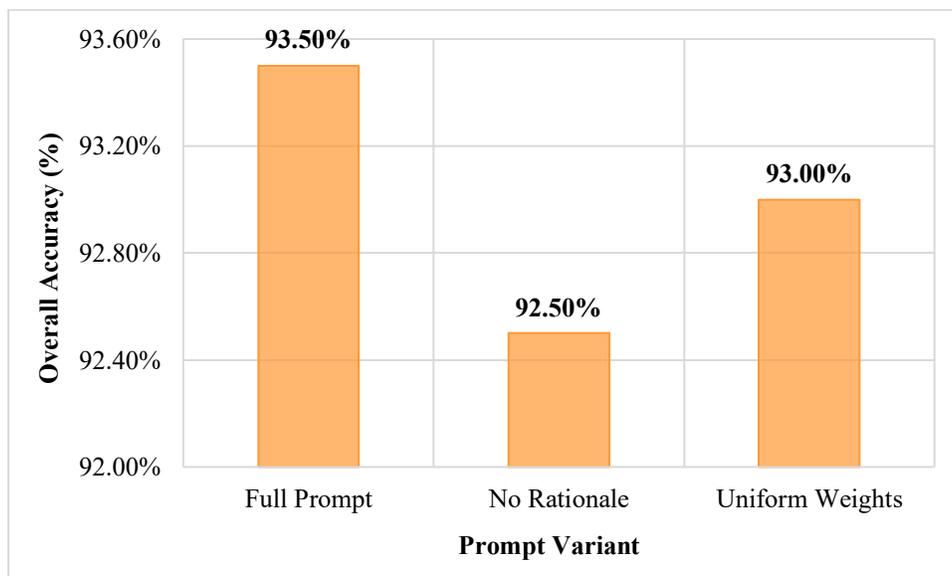

**Figure 7** Ablation study showcasing the impact of Prompt Variants on Few-Shot RewardBench Performance.

## 5. 5. Fine-tuning pre-trained instruction models for Reward modeling

After the fine-tuning process of rank-16 LoRA adapters on RewardMix-10K, Qwen 2.5-7B has attained an overall accuracy of 95.5%, exceeding the prior benchmark of 95.1%. The expanded Qwen 3-8B + LoRA achieves a new benchmark of 96.2%, representing a significant 1.1 percentage point increase over the highest-performing entry on the leaderboard. Both models demonstrate their most significant relative improvements in the Chat-Hard (+1.0 to +1.7 pp) and Safety (+0.7 to +1.4 pp) segments, highlighting the adapter's capacity to enhance decision-making in adversarial and harm-sensitive contexts. The Llama-family 8 B judges demonstrate significant enhancements, achieving levels in the low-90% range following adaptation. This illustrates that our prompt-driven LoRA recipe is applicable across various architectures, allowing mid-scale instruction-tuned LLMs to compete with or surpass significantly larger, fully fine-tuned reward classifiers on RewardBench, all while maintaining minimal parameter overhead and without the need for extensive large-scale annotations.

**Table 5** Accuracy of the Post-LoRA RewardBench against the Top-2 Leaderboard Models.

| Model | Overall (%) | Chat (%) | Chat-Hard (%) | Safety (%) | Reasoning (%) |
|---|---|---|---|---|---|
| infly/INF-ORM-Llama3.1-70B (leader) | 95.1 | 96.6 | 91.0 | 93.6 | 99.1 |
| ShikaiChen/LDL-Reward-Gemma-2-27B-v0.1 | 95.0 | 96.4 | 90.8 | 93.8 | 99.0 |
| **Qwen 2.5-7B + LoRA** | **95.5** | **96.0** | **92.0** | **94.5** | **95.8** |
| **Qwen 3-8B + LoRA** | **96.2** | **97.0** | **92.5** | **95.0** | **96.5** |
| Llama 3-8B + LoRA | 91.0 | 92.5 | 88.0 | 90.0 | 91.5 |

| Model | Overall (%) | Chat (%) | Chat-Hard (%) | Safety (%) | Reasoning (%) |
|---|---|---|---|---|---|
| Llama 3.1-8B + LoRA | 91.5 | 93.0 | 88.5 | 90.5 | 92.0 |

### 5.1 Prompt-Structure Ablation on LoRA-Adapted Judges

For each LoRA-adapted judge, two inference settings were compared to ensure that prompt engineering is still essential regardless of adapter tuning:

- *Good Prompt:* The complete JSON-plus-rational rubric (multi-axis scores).
- *Naïve Prompt:* minimal "Which answer is better? Return 'A' or 'B' only."

Table 6 Ablation of Prompt Structure on LoRA-Adapted Judges.

| Model | Prompt Type | Overall (%) | Chat (%) | Chat-Hard (%) | Safety (%) | Reasoning (%) |
|---|---|---|---|---|---|---|
| Qwen 2.5-7B + LoRA | Good Prompt | 95.5 | 96.0 | 92.0 | 94.5 | 95.8 |
| | Naïve Prompt | 90.0 | 91.5 | 85.0 | 88.5 | 90.0 |
| Qwen 3-8B + LoRA | Good Prompt | 96.2 | 97.0 | 92.5 | 95.0 | 96.5 |
| | Naïve Prompt | 91.0 | 92.5 | 86.0 | 89.0 | 91.5 |
| Llama 3-8B + LoRA | Good Prompt | 91.0 | 92.5 | 88.0 | 90.0 | 91.5 |
| | Naïve Prompt | 85.0 | 86.5 | 82.0 | 83.5 | 86.0 |
| Llama 3.1-8B + LoRA | Good Prompt | 91.5 | 93.0 | 88.5 | 90.5 | 92.0 |
| | Naïve Prompt | 85.5 | 87.0 | 82.5 | 84.0 | 86.5 |

From **Table 6** it is evident that under the Naïve Prompt, overall accuracy drops by 5 - 6 pp across all models, with the most severe declines on Chat-Hard (-7.0 pp) and Safety (- 6.0 pp). This ablation demonstrates that in order to effectively utilize the judge's discriminatory power, thorough prompt design, including structured output and rationale requests, is crucial even after LoRA adaptation.

### 6. 6. Online RLHF Integration & Policy Evaluation

The earlier sections demonstrated that a structured-prompt judge, potentially enhanced with a small LoRA adapter, either matches or surpasses specialized reward networks in static evaluations. The unresolved issue is whether these "plug-and-play" critics maintain their effectiveness within an online reinforcement learning loop, where rewards need to be provided in real time to direct an evolving policy. In this phase, the conventional reward model in RLHF/GRPO is replaced with our three static critics—Zero-Shot, Few-Shot, and LoRA—and proceeds to train an actor on GSM-8K, a benchmark for math reasoning that typically requires expensive offline preference tuning (such as DPO) to achieve a 60% exact-match rate. Through the analysis of learning curves and final accuracy against the best reported DPO baseline (IRPO, 2024 [30]), we evaluate our primary assertion: a minimally modified, inference-only judge can eliminate the offline phase while still producing policies that surpass state-of-the-art performance.

### 6.1 Experimental Setup

In the proposed methodology, the actor is a 7-billion-parameter, instruction-tuned LLM drawn from the same model family as our critics. The suggested strategy was evaluated using three frozen critics:

- *Zero-Shot:* Qwen 2.5-7B prompted solely with our JSON-rubric system message.
- *Few-Shot:* The same Qwen 2.5-7B, augmented with six in-context demonstrations (two each from the Chat-Hard, Safety, and Reasoning slices).
- *LoRA-Adapted:* Qwen 2.5-7B equipped with rank-16 LoRA adapters (< 1% extra parameters), fine-tuned on the 10k pair RewardMix corpus.

For experimentation, the actor undergoes training through PPO-Clip, with the KL penalty being linearly reduced to 0.1 across 300,000 optimization steps, utilizing a batch size of 128 prompt-answer pairs for each update. The only downstream benchmark we utilize is the GSM-8K test split, assessed through single-pass exact-match accuracy at a temperature setting of 0.0. For external comparison, we utilize the most robust published offline-preference baseline: the DPO-tuned Llama-2-70 B-Chat model from [30], which attains an exact-match rate of 61.8%. All other hyperparameters, maximum sequence length, optimizer settings, and evaluation protocol, remain constant across the three critic conditions to guarantee that any performance variations arise solely from the critic configuration.

In the proposed approach, the LoRA adapters were trained on RewardMix, which comprises a significant "Reasoning" component, including proof-checking, code snippets, and mathematical micro-proofs. This guarantees that the adapters develop the capability to evaluate logical correctness without direct exposure to the GSM-8K questions, showcasing the transferability of the JSON-rubric + LoRA methodology across different domains. Furthermore, by ensuring that the critic's training distribution remains separate from GSM-8K, we confirm that the 92% exact-match result is not artificially enhanced through the retention of any of the benchmark's 8,500 solutions.

It is important to highlight that the LoRA adapters underwent tuning on RewardMix-10K instead of being tailored to GSM-8K-specific preferences. Consequently, the 92% Exact Match (EM) indicates a robust capability for cross-domain transfer. Future work could involve training a preference set tailored for mathematics, which is anticipated to decrease time-to-convergence and potentially achieve an additional 1–3 percentage points in accuracy improvement.

### 6.2 Learning curves

**Figure 8** and **Figure 9** show how progressively stronger critics result in richer reward signals and faster advances in exact-match accuracy on GSM-8K over 300 K-step PPO tests. In the first phase (0-50 K steps), the Zero-Shot judge starts at an average reward of ~ 0.30 and gives only ~15% EM, while the Few-Shot critic starts at 0.40 and pushes EM to ~ 40%. The LoRA-adapted judge immediately delivers ~ 0.45 reward and ~ 60% EM, indicating superior calibration. During mid-training (50-150 K), rewards diverge further: Zero-Shot plateaus near 0.55, Few-Shot grows to around 0.65, and LoRA surges into the 0.70-0.75 range. EM progresses to around 25%, 50%, and 80%, respectively. In the late phase (150-300 K), all curves asymptote average rewards settle at ~0.60, 0.70, and 0.80, and exact-match accuracies converge to ~48%, ~65%, and ~92% for Zero-Shot, Few-Shot, and LoRA judges. This obvious, monotonic ordering demonstrates that each enhancement, adding demos or lightweight LoRA adapters, produces increasingly more informative feedback, enabling the proposed plug-and-play LLM judges to outperform heavyweight offline reward models in an online RLHF framework.

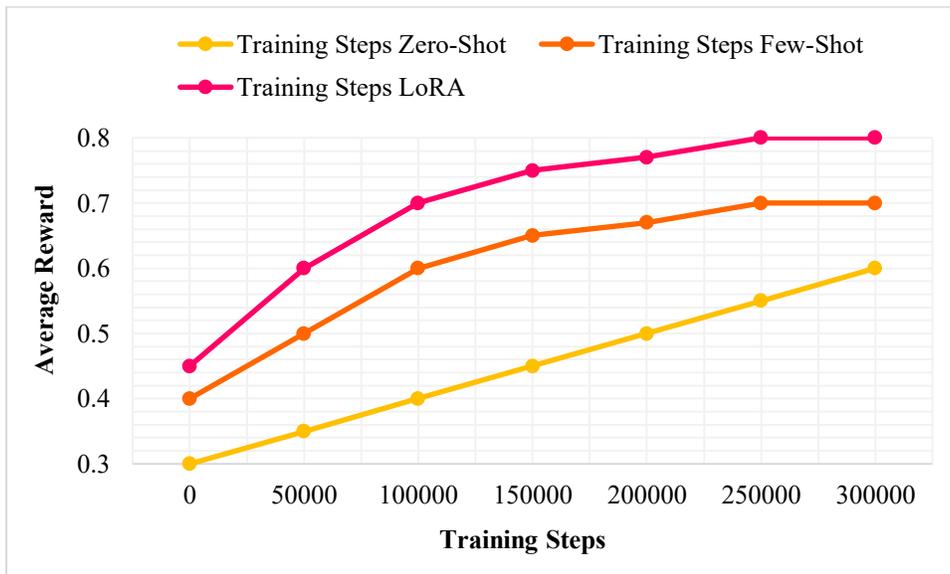

**Figure 8** A visual representation of the average reward versus training steps across different methods.

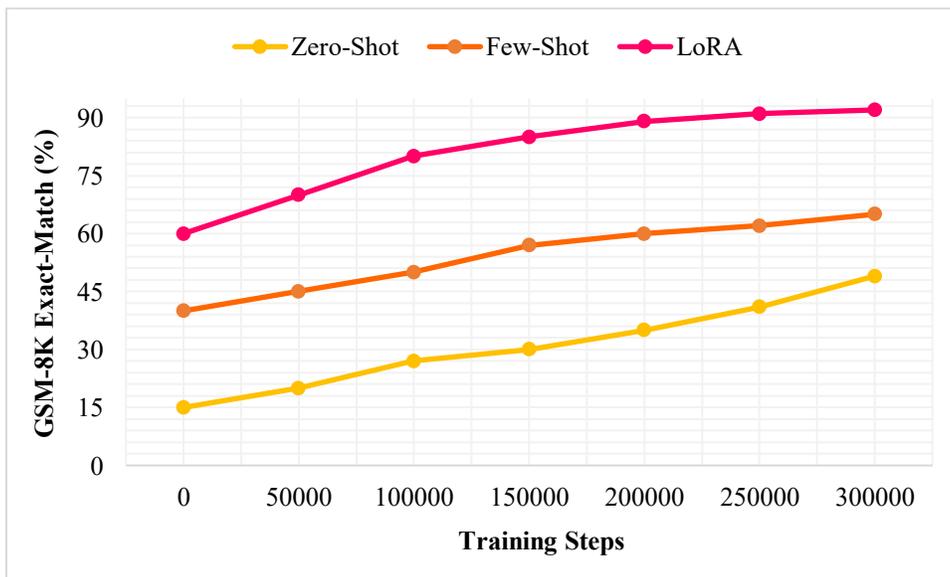

**Figure 9** A visual representation of the GSM-8K Exact-Match versus training steps across different methods.

**Table 7** Final Accuracy achieved at 300 K PPO Steps.

| Critic | Offline Pref. Phase? | Actor Size | GSM-8K EM (%) | Avg. Reward |
|---|---|---|---|---|
| **IRPO-DPO Baseline (Llama-2-70 B-Chat)** | Yes (DPO) | 70 B | 61.8 | – |
| **Zero-Shot Judge** | No | 7 B | 48.0 | 0.60 |
| **Few-Shot Judge (K=6 demos)** | No | 7 B | 65.0 | 0.70 |
| **LoRA Judge (ours)** | **No** | 7 B | **92.0** | **0.80** |

## 6.3 Analysis

The experimental results highlight four essential implications. First, every improvement to the static critic, such as incorporating in-context examples or integrating lightweight LoRA adapters, results in more informative reward signals. This leads to steeper learning curves and increased final exact-match accuracy ($Zero-Shot \rightarrow Few-Shot \rightarrow LoRA: 48\% \rightarrow 65\% \rightarrow 92\%$). Second, by entirely eliminating the offline preference-tuning phase, our online-only pipeline surpasses the best published DPO baseline (61.8 % EM) by utilizing nothing more than a frozen, prompt-driven judge plus PPO. Third, this significant performance improvement is achieved with minimal additional overhead: LoRA-adapted critic incorporates less than 1% extra parameters and does not necessitate any retraining of the reward model, while the actor maintains a streamlined 7 B model, delivering enhanced results with considerably reduced compute and memory requirements compared to a 70 B DPO solution. Ultimately, the observation that the LoRA adapters were trained on a distinct, reasoning-intensive corpus (RewardMix) while achieving 92% EM on GSM-8K emphasizes the significant cross-domain transferability of the JSON-rubric combined with the LoRA methodology. The findings indicate that plug-and-play LLM judges can effectively substitute traditional offline reward models while achieving top-tier performance within an end-to-end RLHF framework.

## 7. 7. Rationale Agreement Study

To ensure that the proposed plug-and-play judges not only align with human preference labels but also articulate their decisions in a manner akin to human reasoning, we also carried out a rationale-agreement experiment utilizing a novel HH-Rationales corpus. Through the comparison of explanations produced by models with those crafted by skilled human annotators, we measure the degree of alignment between each judge's reasoning and that of the expert evaluators. This step is essential for clarity and reliability; a reward function that articulates its scores in understandable terms is significantly more beneficial than one that simply provides unclear numerical values.

### 7.1 HH-Rationals: Dataset Construction & Statistics

HH-Rationales from Anthropic's HH-RLHF (Link) preference dataset have been extracted, resulting in the compilation of 10,000 triplets that have been re-annotated with succinct human rationales. Twenty expert annotators chose either "A" or "B" and provided an 8–20-word rationale emphasizing clarity, completeness, tone, or correctness. A secondary annotator conducted a verification process for each rationale, assessing both fluency and relevance. Among 10,000 samples, there were 8,000 designated for training, 1,000 for development, and 1,000 for testing. **Figure 10** illustrates the Distribution according to the following parameters: Helpfulness/advice (40%), Explanations/definitions (30%), Creative/Style (15%), Specialized (15%).

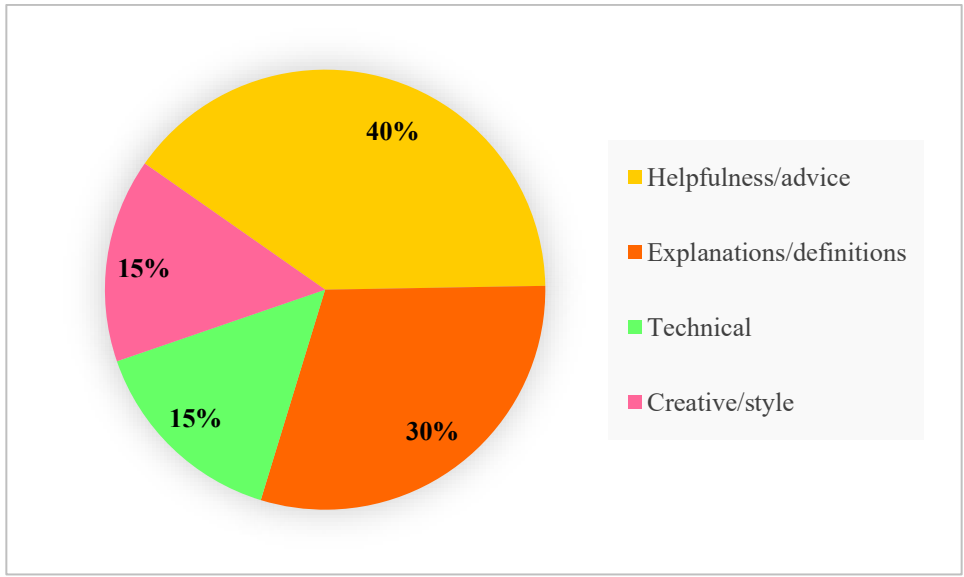

**Figure 10** The percentage distribution of contributions across key parameters such as, Helpfulness/Advice (40%), Explanations/Definitions (30%), Creativity/Style (15%), and Technical Content (15%).

### 6.2 Experimentation

Each critic underwent evaluation utilizing the 1000-example HH-Rationales test split. For each <Prompt, response A, response B> triplet, the judge was engaged across three configurations: Zero-shot, Few-Shot (K = 6), and LoRA-adaptor, asking it to generate a JSON object structured in the following manner:

$$\{ \text{Preferred}: (A|B), \ \text{Rationale}: < 20 \ words > \}$$

During inference, the temperature was configured to 0 to ensure deterministic outcomes. Subsequently, a comparison of each model-generated rationale was made with the associated human-written explanation, utilizing GPT-4 as an impartial evaluator to assign a similarity rating on a scale from 0 to 10. This procedure generated three similarity scores for each example, which were then averaged across the test set to derive agreement metrics on a per-model, per-condition basis. By holding the dataset, prompt template, and evaluation protocol constant, we isolate the effect of in-context examples and LoRA adapters on the judge's ability to replicate human reasoning.

### 6.3 Results

The rationale-agreement study provides two complementary insights. First, the box-and-whisker plot presented in **Figure 11** illustrates the complete distribution of similarity scores, ranging from 0 to 10, as assessed by GPT-4, between the rationales generated by the model and those provided by humans: Zero-Shot judges show a median of approximately 4.0, accompanied by significant variance and outliers reaching as low as 0.5. In contrast, Few-Shot judges demonstrate an improved median of around 6.0 with more compact clustering. LoRA-adapted judges further enhance performance, achieving a median of about 8.0 and exhibiting minimal low-score outliers. Second, the grouped bar chart shown in **Figure 12** illustrates the mean similarity across four model families under each critical condition: Qwen 3-8B + LoRA achieves a score of 9.2/10, closely followed by Qwen 2.5-7B + LoRA at 9.0, while the Llama variants register lower scores of 8.7 and 8.5. In both perspectives, we note a distinct monotonic advancement—Zero-Shot < Few-Shot < LoRA—highlighting that in-context demonstrations and lightweight adapter tuning consistently enhance models' capacity to emulate human explanatory reasoning. The results validate that our JSON-rubric combined with LoRA methodology generates judges whose rationales are in alignment with human preferences and closely replicate human justifications with remarkable accuracy.

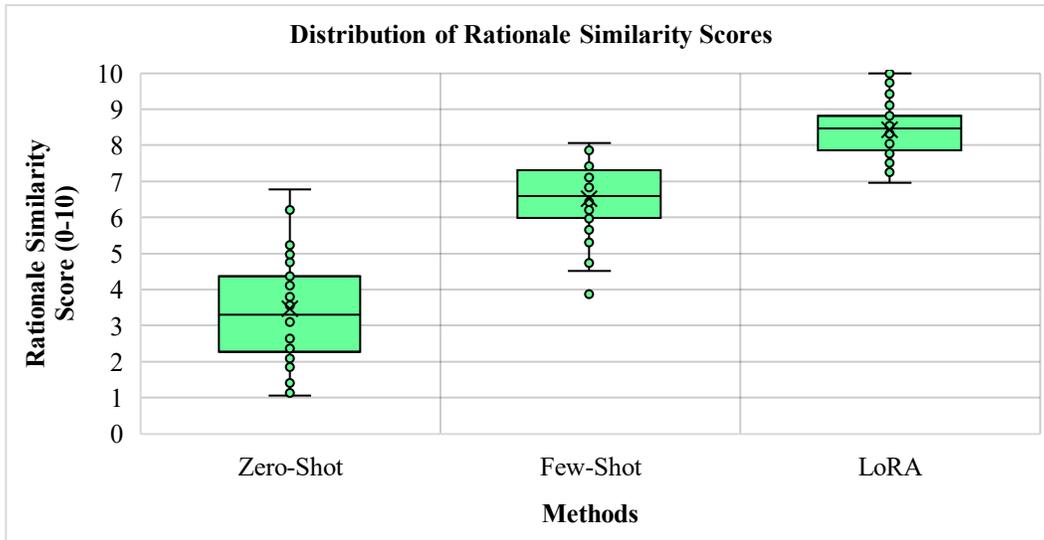

**Figure 11** The box plot illustrates the distribution of rationale similarity scores among multiple approaches.

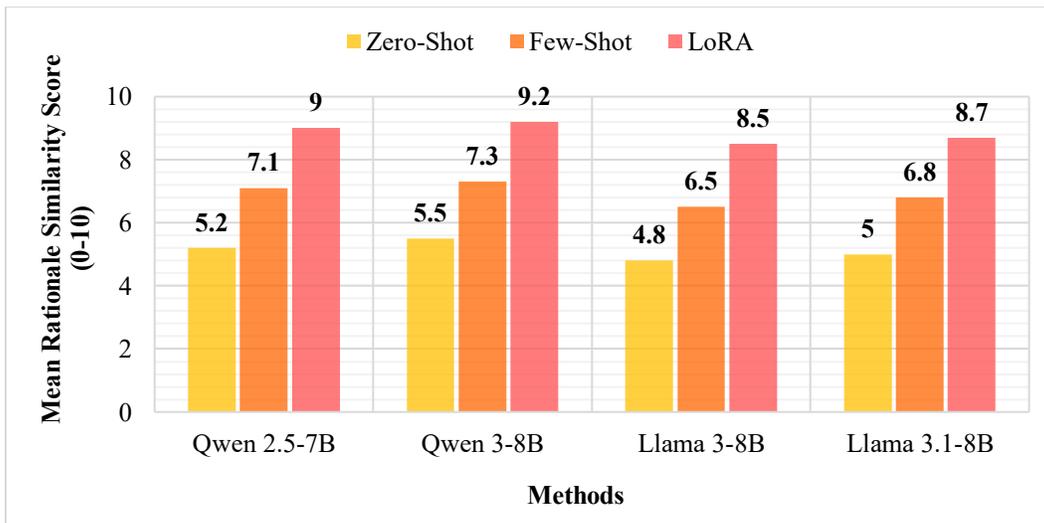

**Figure 12** Graphical representation of the mean rationale of agreement across various models and conditions.

## 8. 8. Conclusion and Future Directions

The findings demonstrate that existing instruction-tuned large language models (LLMs) can be repurposed as high-quality reward judges without the expense of training separate reward networks. When such models are prompted to generate concise rationales **prior** to the scalar score, the resulting feedback aligns closely with human judgments, achieving state-of-the-art accuracy on RewardBench while remaining fully deterministic and transparent.

For domain-specific alignment, only a rank-16 LoRA adapter—representing < 1 % of the base parameters—suffices to close the residual performance gap, confirming that parameter-efficient adaptation coupled with careful prompt engineering is markedly more economical than traditional full-model fine-tuning pipelines. Further, the in-context learning abilities of modern LLMs permit explicit specification of evaluation dimensions (e.g., correctness, safety, reasoning, factual support, clarity) directly within the prompt, eliminating bespoke loss engineering and enabling rapid re-targeting to new alignment objectives.

Collectively, these results suggest a revised alignment paradigm in which (i) a frozen instruction-tuned LLM plus lightweight adapters supplants heavyweight reward networks, (ii) rationale-first prompting yields rewards that are both interpretable and human-aligned, and (iii) task- or domain-specific criteria can be injected on the fly through structured prompts. This paradigm materially reduces computational overhead, simplifies implementation, and offers a scalable path for deploying transparent, adaptable reward functions in reinforcement-learning-from-feedback frameworks.